\documentclass[letterpaper, 10 pt, conference]{ieeeconf}

\IEEEoverridecommandlockouts 
\usepackage{amsmath}                      
\usepackage{amssymb}                      
\usepackage{algorithm}                    
\usepackage{algpseudocode}
\usepackage[T1]{fontenc}                  
\usepackage{graphicx}
\usepackage{subcaption}                  
\usepackage[urlcolor=blue,colorlinks=true]{hyperref} 
\usepackage[utf8]{inputenc}               
\usepackage{soul}                         
\usepackage[usenames,dvipsnames]{xcolor}  
\usepackage{multirow}                     
\usepackage{makecell}					  
\usepackage[font=footnotesize]{caption}
\usepackage[letterpaper]{geometry}
\usepackage{fancyhdr}

\fancypagestyle{firstpage}{
  \fancyhf{}
  
  \geometry{
    top=60pt,
    left=48pt,
    right=48pt,
    bottom=43pt
  }
}

\newcommand{\dof}{{\sc dof}}
\newcommand{\dofs}{{\sc dof}s}
\newcommand{\cspace}{\ensuremath{\mathcal{C}_{space}}}

\newcommand{\env}{\mathcal{E}}
\newcommand{\robots}{\mathcal{R}}
\newcommand{\query}{\mathcal{Q}}
\newcommand{\paths}{\mathcal{P}}
\newcommand{\conflict}{\mathcal{C}}

\newcommand{\algorithmicinput}{\textbf{Input:}}
\newcommand{\algorithmicoutput}{\textbf{Output:}}
\newcommand{\INPUT}{\item[\algorithmicinput]}
\newcommand{\OUTPUT}{\item[\algorithmicoutput]}

\algdef{SE}[DOWHILE]{Do}{doWhile}{\algorithmicdo}[1]{\algorithmicwhile\ #1}%

\makeatletter
\renewcommand{\ALG@beginalgorithmic}{\small}
\makeatother

\algrenewcommand\algorithmicindent{1em} 

\begin{document}

	\title{\LARGE \bf
		Experience-based Subproblem Planning for Multi-Robot Motion Planning }
	
	\author{Irving Solis$^{1}$, James Motes$^{2}$, Mike Qin$^{2}$, Marco Morales$^{2,3}$, and
		Nancy M. Amato$^{2}$
		\thanks{$^{1}$Irving Solis is with the Texas A\&M
			University Department of Computer Science and Engineering, College Station, TX,
			77840, USA, and also with the University of Illinois Urbana-Champaign as a visiting scholar.  \{irvingsolis89\}@tamu.edu.}%
		\thanks{$^{2}$James Motes, Mike Qin, Marco Morales and Nancy M. Amato are with the University of Illinois Urbana-Champaign
			Department of Computer Science, Urbana, IL, 61801, USA. \{jmotes2, moralesa, namato\}@illinois.edu. }%
		\thanks{$^{3}$Marco Morales is also with Instituto Tecnológico Autónomo de México (ITAM)
            Department of Computer Science, Ciudad de México, 01080, México. \{marco.morales\}@itam.mx. He is supported in part by Asociación Mexicana de Cultura A.C.}%
	}
	
	\maketitle

	\thispagestyle{empty} 
	\pagestyle{plain}     
        	\markboth{IEEE Robotics and Automation Letters. Preprint Version. Accepted Month, Year}
        {FirstAuthorSurname \MakeLowercase{\textit{et al.}}: ShortTitle} 
	
\begin{abstract}

Multi-robot systems enhance efficiency and productivity across various applications, from manufacturing to surveillance.
While single-robot motion planning has improved by using databases of prior solutions, extending this approach to multi-robot motion planning (MRMP) presents challenges due to the increased complexity and diversity of tasks and configurations.
Recent discrete methods have attempted to address this by focusing on relevant lower-dimensional subproblems, but they are inadequate for complex scenarios like those involving manipulator robots. To overcome this, we propose a novel approach that 
constructs and utilizes databases of solutions for smaller sub-problems. By focusing on interactions between fewer robots, our method reduces the need for exhaustive database growth, allowing for efficient handling of more complex MRMP scenarios. We validate our approach with experiments involving both mobile and manipulator robots, demonstrating significant improvements over existing methods in scalability and planning efficiency. Our contributions include a rapidly constructed database for low-dimensional MRMP problems, a framework for applying these solutions to larger problems, and experimental validation with up to 32 mobile and 16 manipulator robots.

\end{abstract}
		
\section{Introduction}

Multi-robot systems are important in daily life and in various industrial applications. From manufacturing and warehouse management to surveillance and delivery services, multi-robot systems significantly enhance efficiency and productivity. In these environments, the interactions between robots during mobile or manipulation tasks are often repetitive and exhibit common patterns, suggesting that there is significant potential to learn from and reuse previous experiences. Experience-based planning leverages stored solutions from past scenarios to improve efficiency in new planning tasks by reusing relevant knowledge. To date, most of the research on experience-based motion planning has focused on single robot problems, where the planning space is much simpler and more manageable to learn.



In this paper, we propose an innovative experience-based planning approach for large and complex MRMP problems. An analysis from our previous work, ARC~\cite{smqma-arcasbafhmrmp-24}, indicates that in MRMP problems, even with large numbers of robots, most conflicts require the coordination of only a few robots. Thus, instead of attempting to capture a comprehensive set of experiences for an entire team, we focus on the most relevant interactions among smaller groups of robots. These interactions are modeled as lower-dimensional subproblems, which are stored in a compact database. Since these subproblems involve only a few robots, learning their planning spaces requires fewer experiences, resulting in a database that is both smaller and quicker to compute. This approach enables us to efficiently solve these subproblems and, in turn, tackle larger and more complex scenarios involving many robots.


\begin{figure}[t]
\footnotesize
		\centering
		\begin{subfigure}[b]{0.49\linewidth}
			\includegraphics[width=\linewidth]{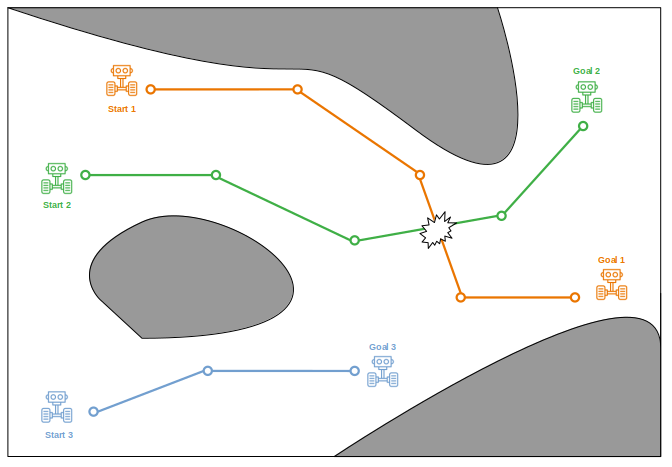}
			\caption{\footnotesize\vspace{0mm}}
		\end{subfigure}
		\begin{subfigure}[b]{0.49\linewidth}
			\includegraphics[width=\linewidth]{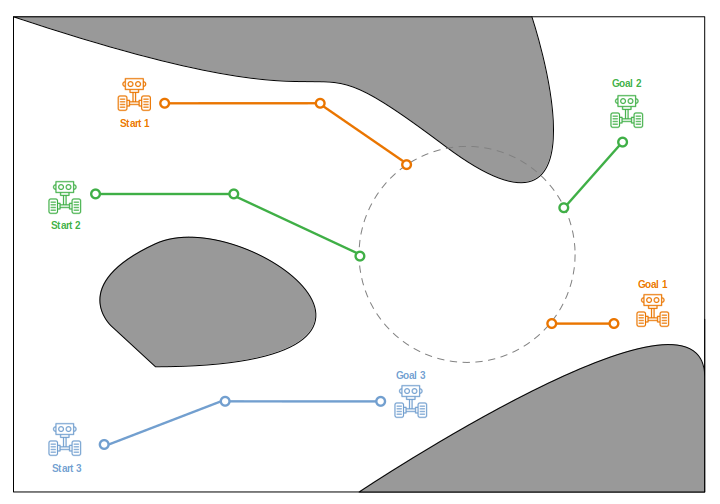}
			\caption{\footnotesize\vspace{0mm}}
		\end{subfigure}
		\begin{subfigure}[b]{0.99\linewidth}
			\includegraphics[width=\linewidth]{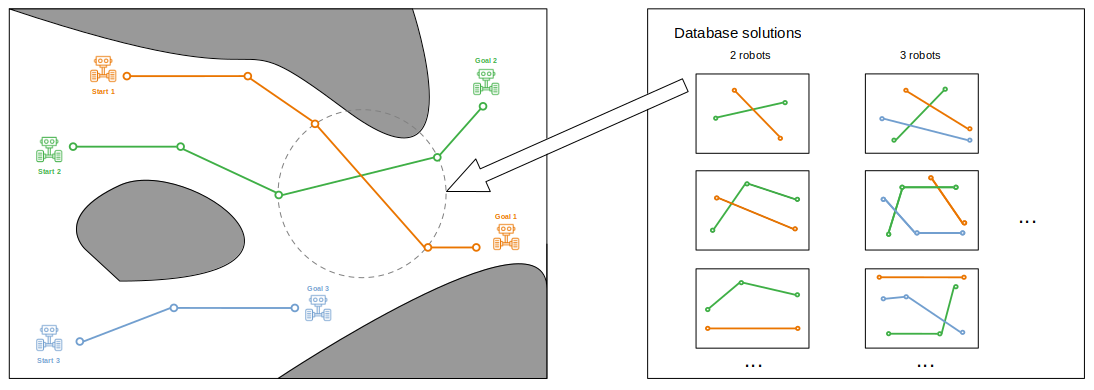}
			\caption{\footnotesize\vspace{0mm}}
		\end{subfigure}
		\caption{{\footnotesize{{A simplified overview of our method: a) Detect a conflict between two robots' paths. b) Define a local subproblem around the conflict. c) Retrieve the best solution from the database and solve the subproblem. }}      
        \vspace{-5mm}}}
		\label{fig:overview}
	\end{figure}
Experience-based planning has proven effective for single-robot motion planning by using databases to store and retrieve meaningful experiences (paths) for solving new problems ~\cite{bag-arppftlfe-12,csmoc-ebpwsrs-15}. While these methods have been adapted to multi-robot scenarios in restricted cases, such as grid-like Multi-Robot Pathfinding (MAPF) ~\cite{hy-dfnomrppudpaosp-20,gy-ehfmrppice-23,mmmah-ebmapfwnc-24}, they primarily address complexity by focusing on lower-dimensional subproblems that involve only the robots in conflict. However, the grid-based, discrete approach of these methods limits their application to high-dimensional, continuous environments like those required for manipulator robots, where flexible and adaptable high-dimensional solutions are essential.


Our work addresses the primary limitation of experience-based planning in MRMP: the exponential growth of the solution database with increasing problem size and complexity. By focusing on lower-dimensional subproblems, we can capture a substantial number of robot interactions and store them efficiently in a compact database. We leverage our previous work ~\cite{smqma-arcasbafhmrmp-24} to identify and create these subproblems within the larger MRMP context. As a result, we can effectively use a smaller, more manageable database to solve these subproblems, allowing us to tackle larger and more complex scenarios.
Additionally, most conflicts require coordination of only a few robots  ~\cite{smqma-arcasbafhmrmp-24}, making solutions for larger groups rarely necessary. This motivates the use of databases for smaller sets of robots.

In this work, we present a novel, feasible, and cost-effective method to build usable databases by leveraging experience-based planning to solve complex MRMP problems with many robots, including both mobile and manipulator robots.
Our method involves constructing and utilizing a database of precomputed solutions for smaller subproblems, such as those involving only 2-4 robots.

We validate our method across various scenarios, including both mobile and manipulator robots arranged in different configurations, with and without obstacles. Our approach does more with less by utilizing a database that is smaller and up to 342 times faster to construct while still providing notable improvements in planning efficiency and scalability compared to state-of-the-art sampling-based and experience-based methods that require much larger databases.

Our contributions are as follows:
    \begin{itemize}
        \item An efficient, fast-constructing database for low-dimensionality MRMP problems. 
        \item A framework to use this low-dimensionality database effectively to leverage experience-based planning in solving large MRMP problems. 
        \item Experimental validation of this proposed method by solving complex problems with up to 32 mobile and 16 manipulator robots by using a smaller database that is up 342 times faster to construct.

        \end{itemize}

\section{Preliminaries and Related Work}

In this section, we define the problem of multi-robot motion planning (MRMP) and outline its challenges. We review traditional planning methods applied to single-robot problems, coupled, decoupled, and hybrid approaches to multi-robot contexts, and experience-based planning techniques for single and multi-robot problems. We also emphasize how our method addresses the limitations of existing approaches.

\subsection{Problem definition}

Motion planning involves finding a feasible path for a robot to move from a start to a target pose within the \textit{configuration space} \cspace that represents all possible states a robot can achieve, taking into account its \textit{degrees of freedom} (\dof), which include variables such as position, orientation, joint angles, and velocity. When this problem is extended to multi-robot systems, known as Multi-Robot Motion Planning (MRMP), the goal is to find feasible paths for multiple robots, each with distinct start and target positions. An MRMP problem is defined by $(\mathcal{E},\mathcal{R},\mathcal{Q})$, where $\mathcal{E}$ represents the environment, $\mathcal{R}$ is the set of robots involved, and $\mathcal{Q}$ denotes the start and goal configurations for each robot.

To solve this, the multi-robot planning space known as the \textit{composite configuration space} $\mathcal{C}_\texttt{composite}$ must be explored, which is the Cartesian product of the individual configuration spaces of all robots, denoted as $\mathcal{C}_\texttt{i}$ for each robot $r_i \in \mathcal{R}$. The exploration of this composite space can be approached in either a coupled or decoupled manner. A valid MRMP solution requires all robots to transition from their start to goal states along conflict-free paths. A conflict occurs when two robots, $r_i$ and $r_j$, interfere with each other at the same timestep $t$ while following their respective paths, denoted as \(\langle c_i, c_j, t \rangle\), where $c_i$ and $c_j$ represent the conflicting configurations.

\subsection{Sampling-based planning}

Sampling-based algorithms enable the solving of high-dimensional motion planning problems by approximating the \cspace\ when an explicit representation becomes infeasible~\cite{c-crmp-88,r-cpmpg-79}. The Probabilistic Roadmap Method (PRM)~\cite{kslo-prpp-96} exemplifies this approach, offering probabilistic completeness instead of exact solutions. PRMs sample the \cspace\ to build roadmaps, where vertices represent valid states and edges represent feasible transitions, facilitating pathfinding through roadmap queries.

\subsection{Multi-robot motion planning methods}

\subsubsection{Coupled Methods}
Coupled approaches directly explore the composite space to find a path from a start to a goal configuration for all robots, often using sampling-based methods (e.g., PRM~\cite{kslo-prpp-96}, RRT~\cite{l-rrtntpp-1998}) to achieve probabilistic completeness and high coordination. A valid configuration within this composite space ensures that no robot collides with obstacles or other robots. Coupled methods naturally avoid such conflicts as they only transition safely between valid configurations. These methods can solve complex scenarios where robots require precise coordination but are limited to small teams due to the exponential growth of the composite space.

\subsubsection{Decoupled Methods}
Decoupled methods plan paths for each robot separately by exploring individual robot configuration spaces using sampling-based techniques. Although these methods simplify planning by not accounting for the degrees of freedom (DOFs) of other robots, it is essential to detect and resolve potential conflicts between paths to achieve a valid solution. This requires additional strategies, such as prioritized planning~\cite{rl-cmrppbhpa-06,bo-pmpmr-05,cgg-pocomrwsg-12}, to manage inter-path conflicts.

\subsubsection{Hybrid Methods}
Hybrid methods combine the strengths of coupled and decoupled approaches, typically constructing decoupled representations and applying MAPF techniques from grid world problems~\cite{wc-sefmpp-15,ssfs-cbsfomap-15}. These methods address path conflicts either by expanding the search space (e.g., M*~\cite{wc-sefmpp-15}) or by imposing constraints on individual robot paths (e.g., CBS~\cite{ssfs-cbsfomap-15}) to ensure collision-free planning. The latter has been extended in CBS-MP~\cite{smsm-romrmpucbs-21}, to allow querying individual sampling-based roadmaps to tackle more complex problems. Furthermore, ARC~\cite{smqma-arcasbafhmrmp-24} adapts robot coordination by breaking down conflicts into subproblems, focusing only on the required robots and planning within a relevant region of their composite space.

\subsection{Experience-based planning}


In contrast to traditional sampling-based methods, which plan from scratch, experience-based approaches leverage previous experience by storing them in a database. Instead of generating solutions from scratch, these methods retrieve and adapt stored solutions to solve new problems. 

Methods that address single-robot motion planning in high-dimensional spaces include~\cite{bag-arppftlfe-12}, \textit{Lightning} employs a parallel framework with two synchronized modules: one that generates paths from scratch and another that retrieves and repairs (RR) paths from a pre-existing database. If a retrieved path results invalid, BIRRT~\cite{kl-rrtceaspp-00} is used to repair the invalid sections. After generating a new path, a database manager decides whether to store it, considering factors such as computation time and similarity to existing solutions. Results show that RR outperforms planning from scratch as the database grows, resulting in notable efficiency improvements. Thunder ~\cite{csmoc-ebpwsrs-15}, extend this with .a roadmap spanner to manage database storage more efficiently, reducing both insertion and retrieval times and further improving execution time compared to its predecessor. However, building a useful database requires solving numerous prior problems. Constructing such a large database with many \dofs for more complex problems can become impractical making these methods unsuited for addressing directly MRMP problems involving many robots, where both the number of potential tasks and the size of the composite configuration space grow exponentially.

Recent research in multi-robot pathfinding (MRPF) has addressed these limitations in grid-like problems. Instead of attempting to solve the entire problem and generating a database of solutions for the full multi-robot system, these approaches create a database of solutions for subproblems involving fewer robots. Based on the observation that most conflicts can be resolved in a reduced space, these methods leverage precomputed solutions to resolve conflicts in the main problem. In~\cite{hy-dfnomrppudpaosp-20}, DDM constructs a solution database for all 2 × 3 and 3 × 3 subproblems. It plans individual paths in a decoupled manner, and when conflicts arise, the database is queried to find a matching subproblem, using its solution to repair paths and resolve conflicts. In~\cite{gy-ehfmrppice-23}, DCBS enhances the conflict-based framework~\cite{ssfs-cbsomapf-12,bssf-svofcbsaftmapp-14} by integrating DDM as the primary method for conflict resolution. Finally, in~\cite{mmmah-ebmapfwnc-24}, a more comprehensive database is constructed to handle narrow passage grid-like problems. However, while these methods effectively address the exponential complexity of experience-based planning in multi-robot scenarios, their discrete nature limits their applicability to real-world continuous problems.

In contrast, our proposed method leverages experience-based planning to tackle complex problems involving many mobile and manipulator robots. It constructs a database of lower-dimensional subproblems, learning specific and useful robot interactions, such as conflict resolution between two or more robots. Our method then retrieves and applies efficiently these subproblem solutions efficiently to scenarios involving a larger number of robots.

\section{Method}
\label{method}

Here, we present Experience-based Adaptive Robot Coordination (E-ARC), an experience-based approach for multi-robot motion planning (MRMP). 
E-ARC manages a database of subproblem solutions to address inter-robot conflicts. It enhances the ARC framework by prioritizing the retrieval of solutions from the database before resorting to planning from scratch.

It starts by constructing a database of valid solutions for randomly generated subproblems in isolation. A subproblem $(\env',\robots',\query')$ is a smaller instance contained within the main problem, considering a reduced portion of the environment $\env' \in \env$, a subset of robots $\robots' \in \robots$, and subqueries $\query'$ within $\env'$. Once the database is established, E-ARC follows the ARC framework to detect conflicts as shown in Alg.\ref{alg:1}. For a given MRMP problem instance $(\env,\robots,\query)$, E-ARC starts by generating individual paths using standard sampling-based techniques. It then checks these paths for conflicts, and any conflicts between paths $p_i, p_j$ are used to create local subproblems $(\env',\robots'= r_i\cup r_j,\query')$.

\begin{algorithm}[t!]
    \caption{Experience-based Adaptive Robot Coordination (E-ARC)}\label{alg:ARC}
    \begin{algorithmic}[1]
        \INPUT A MRMP problem with an environment $\env$, a set of robots $\robots$, a set of queries $\query$. A database $\mathcal{DB}$ with $n$ solutions of different dimensionalities.
        \OUTPUT A set of valid paths $\paths$.
        \State $\paths\leftarrow\emptyset$ \label{alg:empty_paths_1}
        \For {each robot $r_i \in \robots$ } \label{alg:initial_paths}
        \State $p_i\leftarrow$\texttt{IndividualPath}($\env,\{r_i\},\{q_i\}$)
        \State $\paths\leftarrow \paths\cup\{p_i\}$
        \EndFor \label{alg:empty_paths_2}
        \State {$\conflict$ = \texttt{FindConflict($\paths$)}} \label{alg:first_conflict}
        \While {$C$ $\neq \emptyset$}
            \State \texttt{$\env'$,$\robots'$,$\query'$ = CreateSubProblem($\conflict,\paths,\env$)} \label{alg:create_subproblem}
            \State \texttt{$\paths'$ = DatabasePlanning($\env', \robots', \query'$)}
             \If {$\paths' = \emptyset$} 
                 \State \texttt{$\paths'$ = TraditionalPlanning($\env', \robots', \query'$)}
            \EndIf
            \If {$\paths' \neq \emptyset$} \Comment{conflict resolved}
                \State \texttt{UpdateSolution($\paths,\paths'$)} 
                \State {$\conflict$ = \texttt{FindConflict($\paths$)}}\label{alg:find_additional_conflict}

           \Else 
                    \State $\paths = C \leftarrow \emptyset$ \Comment{if conflict not resolved,}
                \EndIf \Comment{return empty solution}
			\EndWhile
			\State\Return $\paths$\label{alg:return_solution}			
    \end{algorithmic}
    \label{alg:1}
\end{algorithm}

Similar to Lightning~\cite{bag-arppftlfe-12} and Thunder~\cite{csmoc-ebpwsrs-15}, E-ARC retrieves from the database the \textit{k} closest solutions of the same dimensionality as the given subproblem. These solutions are then checked for collisions to identify a valid one that can resolve the conflict. If no valid solution is found, E-ARC either repairs the solution with the fewest collisions or resorts to the hierarchy of sampling-based methods, as in the standard ARC framework, to obtain a subproblem solution. Whether the solution comes from the database or from scratch, it resolves the conflict and the updated solution is integrated with the initial paths. This process continues until all conflicts are resolved.

If the hierarchy of planning methods fails to find a solution, the local subproblem $(\env',\robots',\query')$ is expanded by extending $\query'$ further from the conflict on $p_i$ and $p_j$, with $\env'$ adjusted accordingly. E-ARC then attempts to solve this expanded subproblem, first by querying the database and, if necessary, by using the hierarchy of traditional methods.

Regardless, whether the solution is retrieved from the database or planned from scratch, the resulting local solution $\paths'$ for $(\env',\robots',\query')$ resolves the conflict in $p_i, p_j$. As in regular ARC, if $\mathcal{P}'$ conflicts with another subproblem solution, a new local subproblem is introduced to account for all conflicting robots.

\subsection{Database Construction}

E-ARC stores subproblem solutions across various dimensionalities. The database construction involves running random subproblems of lower dimensionalities in isolation and storing their solutions. While a subproblem $(\env',\robots',\query')$ inherits the general structure of a regular MRMP problem, it must be created under the same conditions as conflicts that are turned into subproblems, occurring within reduced regions of the multi-robot planning space. A crucial requirement is that the Euclidean distance between the start and goal of $\query'$ must be relatively short. Similarly, $\query'$ is used to define a local region $\env'$ to ensure that subproblem solutions are computed within a reduced region of the planning space. Here, we discuss how we build such databases for mobile and manipulator robots. 

\begin{figure}[h!]
\centering
\begin{subfigure}[b]{0.43\linewidth}
    \includegraphics[width=\linewidth]{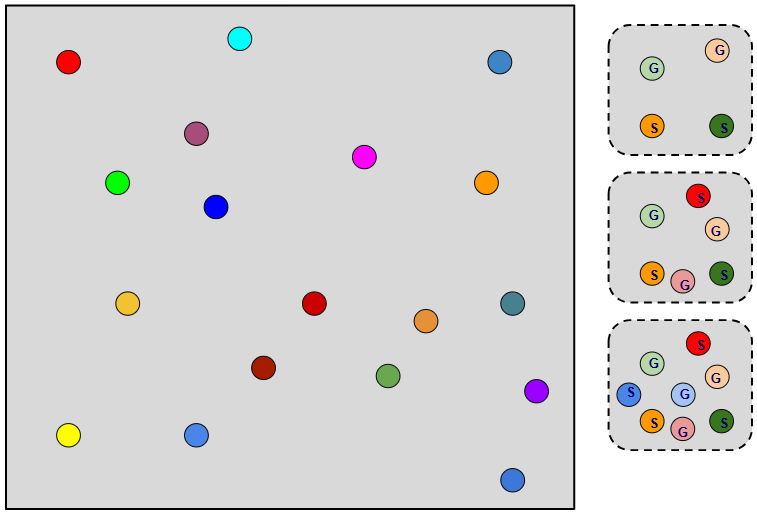}
    \caption{}
    \label{fig:earc-fig5}
\end{subfigure}
\begin{subfigure}[b]{0.55\linewidth}
    \includegraphics[width=\linewidth]{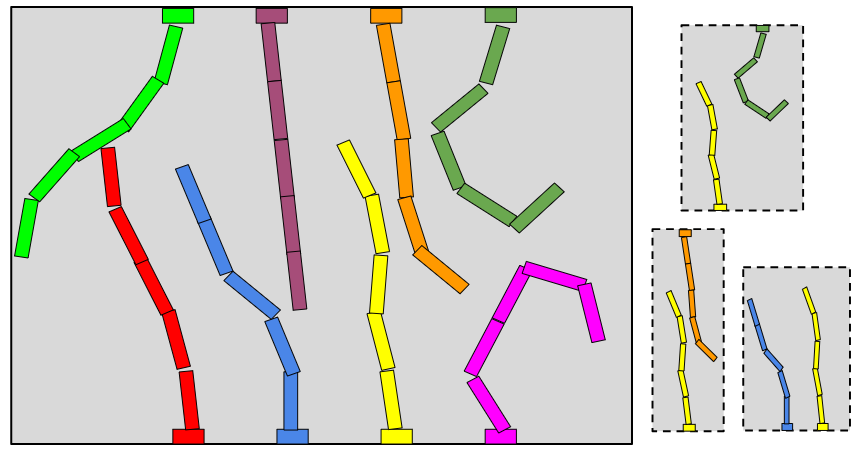}
    \caption{}
    \label{fig:earc-fig6}
\end{subfigure}

\caption{{Construction of the subproblem solution database: (a) For mobile robots, a reduced boundary is used to generate random subproblems involving 2, 3, and 4 robots. (b) For fixed-base manipulators, random subproblems for 2 robots are created for various arrangements, including horizontal, vertical, and diagonal}}
\label{fig:earc-db-const}
\end{figure}

For mobile robots, ARC~\cite{smqma-arcasbafhmrmp-24} analysis, indicates that conflicts that require the coordination of only a few robots are more frequent than those that require many robots for coordination. Therefore, as shown in Fig.\ref{fig:earc-fig5} we generate \textit{n} low dimensional random subproblems $(\env',\robots',\query')_n$ with $\robots'$ consisting of 2, 3, or 4 robots. We assign random valid values to $\query'$ and define a reduced $\env'$ that is just large enough to contain the robots and provide space for computing the solutions. This approach ensures the production of solutions that are useful for conflict resolution. Subproblems are solved using sampling-based techniques within the corresponding composite planning space. 

For fixed-based manipulator robots, additional considerations are necessary, as conflicts between robots may involve different planning spaces. A preliminary step in constructing the database is to classify the various subgroups of robots (Fig.\ref{fig:earc-fig6}) according to their respective planning spaces. This is achieved by recording the relative transformations of the manipulators' bases. We then ensure that only database solutions matching the same relative base transformation are considered. Following this classification, we proceed similarly to the mobile robots by constructing our database by solving \textit{ n} valid 2-robot subproblems for the different subgroups corresponding to distinct planning spaces. Subproblems are also solved using sampling-based techniques within the corresponding composite planning space.

\subsection{Subproblem Creation}

One of ARC's key contributions is its ability to create subproblems derived from conflicts, focusing only on the necessary robots, a relevant portion of the environment, and localized queries. E-ARC builds on this idea by leveraging the subproblem creation process to utilize its database of subproblem solutions to effectively resolve conflicts. This process is explained in more detail below.

Given a set of paths $P$ and a conflict $(c_i,c_j,t)$ between $p_{R_i}$ and $p_{R_j} \in P$, a local subproblem $(\env',\robots',\query')$ is defined around the conflict, where $\robots' = R_i \cup R_j$ includes only the robots involved. The local query $\query'$ sets the start and goal configurations for each robot, based on their positions within a time window before and after the conflict, specifically from $t-\texttt{window}$ to $t+\texttt{window}$.

The local region $\env'$ is defined by a boundary within the \cspace\ around $\query'$, allowing planning methods to focus the composite space search on this localized area around the conflict.

If no solution is found for $(\env',\robots',\query')$, the subproblem is adapted by expanding $\query'$ and $\env'$. This expansion involves pushing the query points further along the paths, thereby enlarging the local environment, and is repeated until a solution is found or until $\env' = \env$ and $\query' = \query$, at which point the method concludes with no solution.

If additional robots are needed to resolve the conflict, E-ARC expands $\robots'$ to include all involved robots, ensuring that the solution remains feasible and prevents situations where resolving one conflict creates a new one with a previous resolution.

\subsection{Database Retrieval and Subproblem Planning}

Each time E-ARC encounters a new conflict, it converts the conflict into a subproblem to resolve it, with the initial approach being to utilize its database, as shown in Alg.\ref{alg:2}. E-ARC retrieves a set \textit{k} of the closest candidate solutions from the database that match the same dimensionality (and the same relative base transformation, in the case of manipulator robots). "Closest" refers to the shortest Euclidean distance between the start/goal of the subproblem and the database solutions' start/goal. Before querying the database, all solutions must be transformed to align with the context location of the conflict. For mobile robots, this involves translating the database solutions to the geometric point where the conflict occurred. For manipulator robots, only solutions with the same relative base transformation are considered. If the transformation matches positively, the solution can be directly applied; if it matches negatively, indicating the roles of the robots are inverted, the solution is transformed accordingly to assign the correct values to the appropriate robots.

The set of retrieved paths is validated to ensure they are collision-free. The first step is to check whether each path's start and goal configurations can be connected to the subproblem's start and goal without collisions; if they cannot be connected, the solution is discarded. The remaining paths in the set are then checked for obstacle collisions to find a suitable, collision-free option for resolving the subproblem. If a collision-free solution is found, it is used to solve the subproblem (and thus resolve the conflict) by connecting its start and goal to the subproblem's start and goal.

If no collision-free solution is found in the database, the path with the fewest invalid segments is selected. E-ARC then decides whether to repair the solution or resort to the ARC hierarchy of traditional planning methods, based on the number of invalid segments. This decision is critical, as the cost of repairing a single invalid segment can be nearly equivalent to solving the entire subproblem from scratch, making it inefficient to repair a solution with many invalid segments.

	\begin{algorithm}[t]
		\caption{DatabasePlanning}\label{alg:solve_subproblem}
		\begin{algorithmic}[1]
			\INPUT A subproblem with an environment $\env'$, a set of robots $\robots$, a set of queries $\query'$.
                 A database $\mathcal{DB}$ with $n$ solutions of different dimensionalities.
			\OUTPUT A set of valid local paths $\paths'$.
            \State $\paths'\leftarrow\emptyset$
            \State $\mathcal{DB}\leftarrow$\texttt{TransformPaths($\mathcal{DB},\query'$)}
            \State $\paths_c\leftarrow$\texttt{kClosestPaths($\mathcal{DB},\query'$)}
            \State $p\leftarrow$\texttt{ValidatePaths($\paths_c$)}
            \If {$p.minCollisions = 0$ }
                \State {\texttt{ConnectPath($p,\query'$)}}              
            \ElsIf {$p.minCollisions < maxCollisions$ } 
                \State {\texttt{RepairPath($p$)}}
                \State {\texttt{ConnectPath($p,\query'$)}}
            \Else
                \State \Return $\emptyset$  \Comment{if database fails,}
            \EndIf \Comment{return empty local solution}

		\end{algorithmic}
    \label{alg:2}
	\end{algorithm}

\section{Experiments}

In our experiments, we demonstrate two key points: first, that E-ARC can achieve "more with less" by efficiently leveraging a small, fast-to-construct database of subproblem solutions, in contrast to the high cost and memory demands of a experience-based planner that uses a database for entire MRMP problems. Second, we show E-ARC's improvement over standard ARC~\cite{smqma-arcasbafhmrmp-24}  by utilizing its subproblem database to resolve conflicts, rather than planning from scratch.

\subsection{Experimental setup}

To evaluate our method's performance, we compare E-ARC against two baselines: a full-problem MRMP database built by Lightning~\cite{bag-arppftlfe-12} and the standard ARC, a state-of-the-art traditional planning method that models conflict resolution as local subproblems.

We conduct evaluations in two distinct scenarios: one for mobile robots and another for manipulator robots. Each scenario is tested with and without random obstacles that change their positions in each trial to assess how the methods perform when obstacles are included, with a particular focus on how database-driven methods adapt to changes caused by obstacle collisions. Additionally, we test the scalability of E-ARC by increasing the number of robots, analyzing both the database construction time and the planning time required to solve problems.

The MRMP database is constructed using Lightning's parallel framework, which simultaneously runs two planning-from-scratch versions of RRT-Connect (decoupled and coupled), alongside the retrieve-and-repair process that leverages the library. The first solution found is added to the database, and only solutions to new subproblems are subsequently included to maintain efficient storage. As the number of robots increases in our experiments, we allow the database to grow proportionally to account for the increasing complexity. 
For the E-ARC database, we use a single database for all mobile robot scenarios and another for manipulator robot scenarios. For mobile robots, the database includes solutions to subproblems with 2, 3, and 4 robots, as most conflicts typically involve a limited number of robots. For manipulator robots, we focus only on 2-robot subproblems, as this simplifies the implementation by allowing us to capture and classify their relative fixed-based transformations efficiently. In future work, we aim to explore efficient ways for effectively managing and classifying multi-robot transformations involving more than two fixed-based manipulator robots. We report the size of these databases and the time taken to construct them.

ARC is implemented using a single-robot PRM to find individual paths. Then, DecoupledPRM and CoupledPRM are employed in sequence within a hierarchy of subproblem solvers to resolve conflicts. E-ARC is implemented similarly, with the key difference being that the database of subproblem solutions is prioritized as the first choice in the hierarchy of subproblem solvers.

We conducted 100 random trials for each scenario, providing random valid start and goal positions for all robots. Each trial was allowed a maximum of 1,000 seconds for planning; any trial exceeding this time limit was marked as a failure. We compared the time required by each method to find the first solution and reported the database construction time for methods that use databases. Solution costs were not reported since each trial involved different queries.

\subsection{Scenario I: Mobile Robots}

\


In this scenario, we evaluate problems involving 3-DOF mobile robots in two environments with and without random obstacles.

\begin{figure}[h!]
\centering
\begin{subfigure}[b]{0.49\linewidth}
    \includegraphics[width=\linewidth]{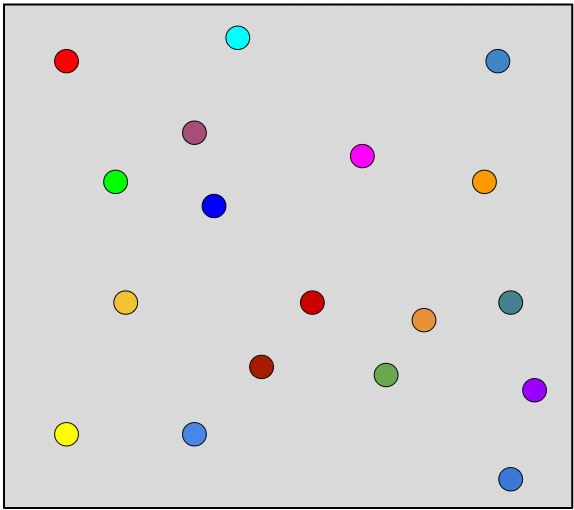}
    \caption{}
    \label{fig:earc-fig1}
\end{subfigure}
\begin{subfigure}[b]{0.475\linewidth}
    \includegraphics[width=\linewidth]{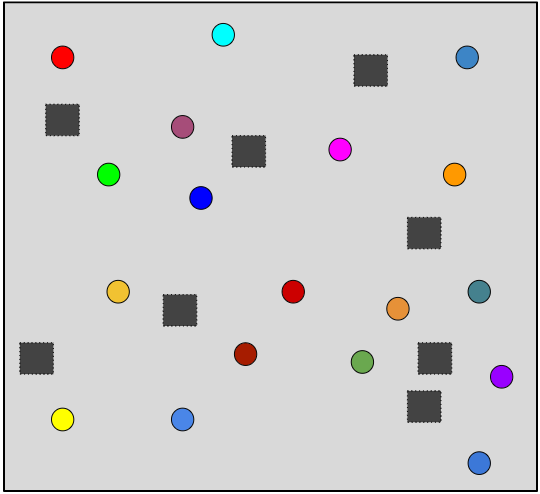}
    \caption{}
    \label{fig:earc-fig2}
\end{subfigure}

\caption{{Multi-mobile robot scenarios: Robots move from random start positions to random goal positions in environments both without (a) and with (b) random obstacles.}}
\label{fig:earc-mobile}
\end{figure}

The first environment (Fig.~\ref{fig:earc-fig1}) is completely empty, without obstacles. This serves as a baseline to analyze the performance of the methods without obstacle collisions. The second environment  (Fig.~\ref{fig:earc-fig2}) introduces a number of random obstacles positioned in different positions for each trial, allowing us to assess how database methods are affected by the need to adapt or repair solutions that were valid for a specific environment but become invalid due to the new positioning of obstacles.

For each of these environments, we test an increasing number of robots: 2, 4, 8, 16, and 32. This approach enables us to analyze the scalability of the database methods in terms of both database construction and the retrieval, validation, and repair of paths from the database.

\subsubsection{Results}

In the scenario without obstacles, as shown in Fig. \ref{fig:plot1}, E-ARC clearly demonstrates greater efficiency compared to Lightning, particularly as the number of robots increases. Lightning is significantly impacted by the growth in robot numbers due to several factors. First, as the database size expands with more entries, the time required for querying increases substantially. Second, the complexity of the database entries themselves also grows; for instance, representing the state of 16 robots is much more computationally expensive than representing the state of just 2 robots, making it more costly to check each path when searching for the closest paths. Third, with a larger number of robots, a greater number of "k-nearest" paths must be retrieved, as it becomes less likely to find a path that can be connected collision-free. Fourth, validating these paths becomes increasingly expensive, as determining the number of collisions in higher-dimensional paths requires more intensive computations. Consequently, these factors contribute to a substantial increase in both database construction time (from 18.79 seconds for 2 robots to 2,607.2 seconds for 16 robots) and planning time (from 0.037 seconds for 2 robots to 5.75 seconds for 16 robots). Due to these escalating complexities, Lightning was unable to solve any of the 100 trials for the scenario involving 32 robots, highlighting its limitations in handling larger-scale problems.

In contrast, E-ARC demonstrates its capability to "do more with less" by constructing a single, compact database that is reused across all scenarios, regardless of the number of robots. This results in a consistent and much lower database construction time of 7.42 seconds across all cases, reflecting E-ARC's scalability and efficiency. Additionally, E-ARC significantly improves upon ARC by utilizing its precomputed subproblem solutions database as the primary method for conflict resolution, leading to faster planning times in all scenarios. For example, for 32 robots, E-ARC achieves an average planning time of 3.05 seconds, compared to ARC's 5.24 seconds. These results underscore E-ARC’s effectiveness in managing increasing numbers of robots and its superior planning efficiency over both Lightning and ARC, making it a robust and scalable solution for complex multi-robot motion planning problems.

When introducing random obstacles into the environment, as shown in Fig.  \ref{fig:plot2}, the complexity of the problem increases further, presenting additional challenges for the planning methods. This scenario proves especially difficult for Lightning, which must find solutions within a database that now has to account for these varying conditions. As the environment becomes more constrained due to the presence of obstacles, Lightning's ability to find collision-free paths diminishes significantly. This results in Lightning being able to solve only problems with 2 and 4 robots, and even then, it completes just 82 out of 100 trials for 2 robots and only 22 out of 100 trials for 4 robots. In contrast, E-ARC maintains its effectiveness despite these variations in the environment. By leveraging a database composed of small subproblem solutions, E-ARC ensures that most of its solutions remain applicable even when obstacles are introduced. Even as the number of robots increases, E-ARC maintains relatively low planning times (e.g., 5.53 seconds for 32 robots) compared to ARC (9.1 seconds for 32 robots), highlighting its efficiency in resolving conflicts and planning in varied environments. These results emphasize E-ARC's robustness and scalability, making it a superior solution for multi-robot motion planning in both static and variable environments.



\subsection{Scenario II: Mobile Robots}

In this scenario, we examine problems involving 5-DOF planar manipulator robots, which require a higher level of coordination to resolve conflicts compared to mobile robots. Here, we aim to evaluate the computational complexity of constructing databases capable of handling such high-complexity problems. As in the previous scenario, we analyze two different obstacle densities, each tested with an increasing number of robots.

\begin{figure}[h!]
\centering
\begin{subfigure}[b]{0.49\linewidth}
    \includegraphics[width=\linewidth]{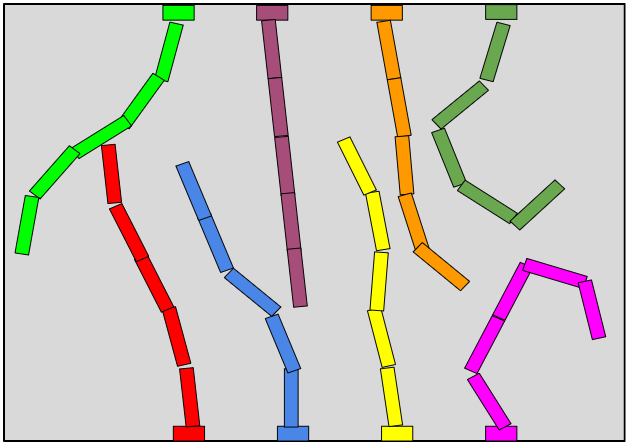}
    \caption{}
    \label{fig:earc-fig3}
\end{subfigure}
\begin{subfigure}[b]{0.49\linewidth}
    \includegraphics[width=\linewidth]{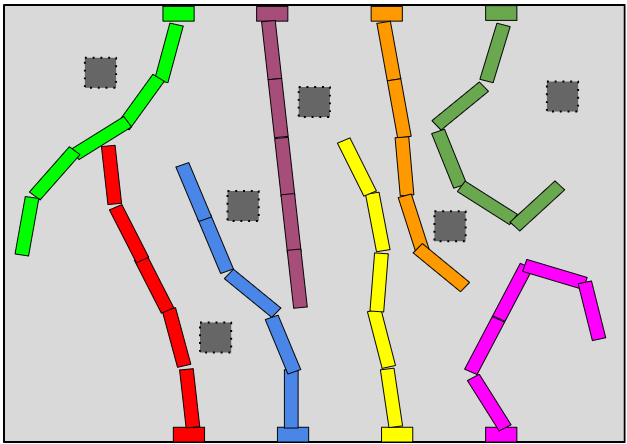}
    \caption{}
    \label{fig:earc-fig4}
\end{subfigure}

\caption{{Multi-manipulator scenarios: Robots move from random start positions to random goal positions in environments both with (b) and without (a) varying obstacles.}}
\label{fig:earc-manipulator}
\end{figure}

\begin{figure*}[h!]
		\centering
        \includegraphics[width=\linewidth]{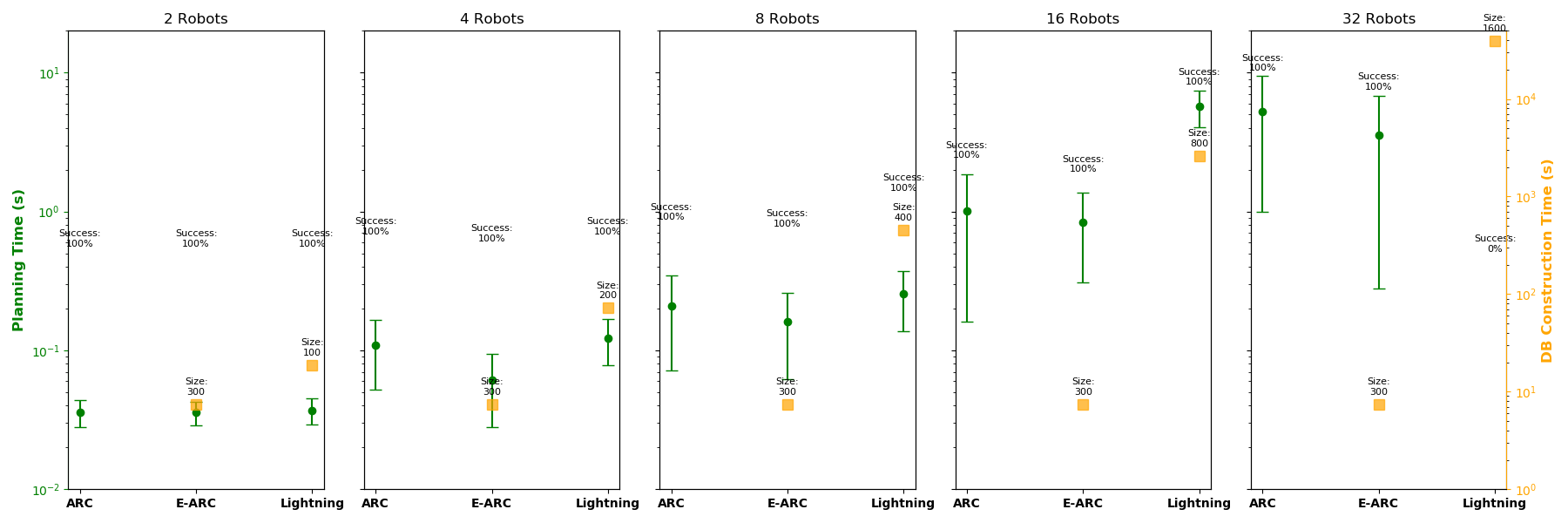}
        \caption{\footnotesize Results for the scenario of mobile robots without obstacles \vspace{-3mm}}
		\label{fig:plot1}

	\end{figure*}

\begin{figure*}[h!]
		\centering
        \includegraphics[width=\linewidth]{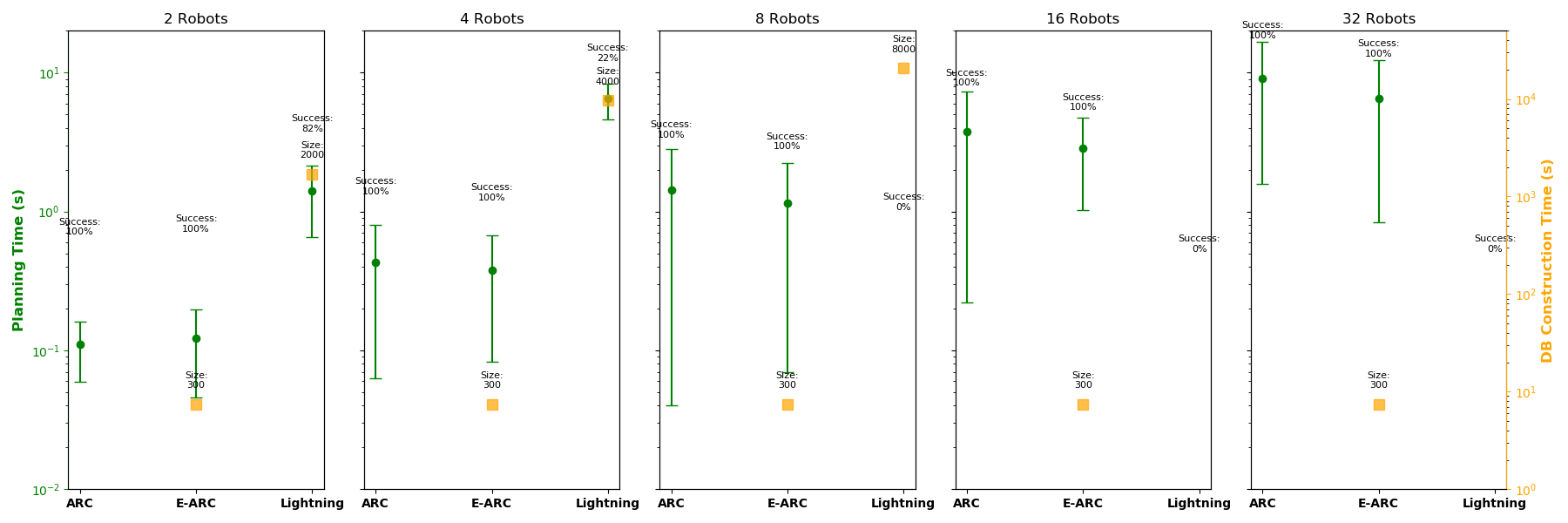}
        \caption{\footnotesize Results for the scenario of mobile robots with random obstacles \vspace{-3mm}}
		\label{fig:plot2}

	\end{figure*}

The manipulator robots are arranged in pairs with their fixed bases positioned opposite each other. Depending on the randomly generated valid queries, each trial may involve conflicts between neighboring robots. Similar to the previous scenario, the first environment (Fig.~\ref{fig:earc-fig3}) is completely empty, while the second environment (Fig.~\ref{fig:earc-fig4}) includes a number of random obstacles.

\subsubsection{Results}


    \begin{figure*}[h!]
		\centering
        \includegraphics[width=0.8\linewidth]{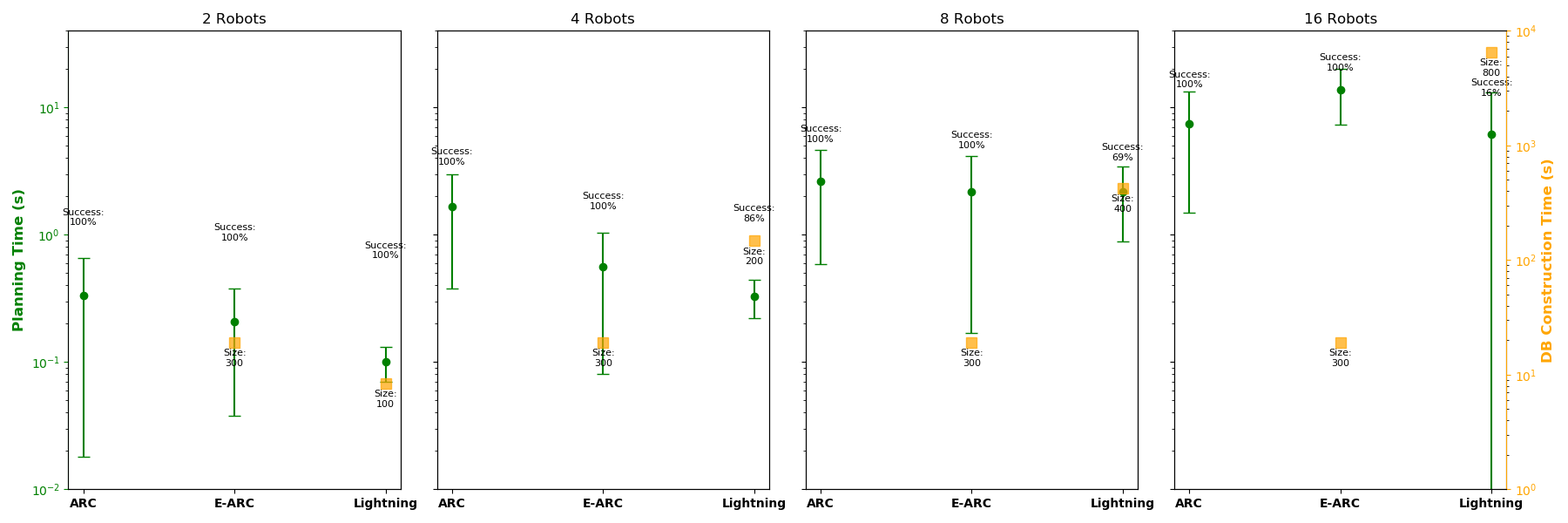}
        \caption{\footnotesize Results for the scenario of manipulator robots without obstacles \vspace{-3mm}}
		\label{fig:plot3}

	\end{figure*}
    \begin{figure*}[h!]
		\centering
        \includegraphics[width=0.8\linewidth]{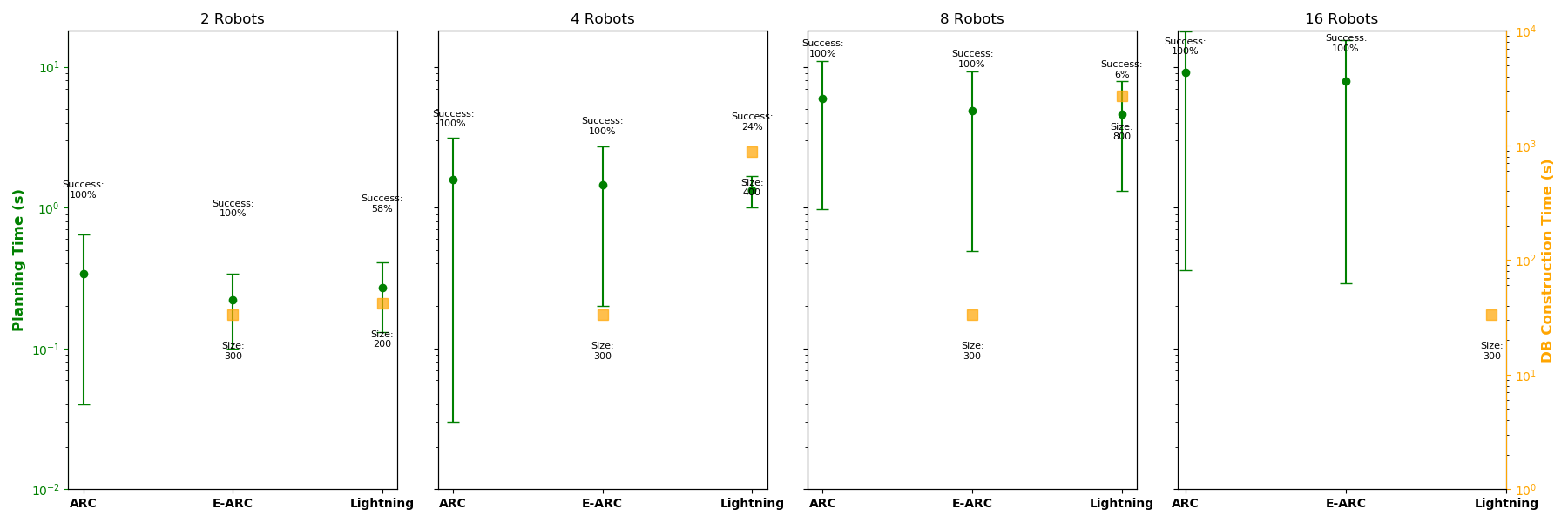}
        \caption{\footnotesize Results for the scenario of manipulator robots with random obstacles \vspace{-3mm}}
		\label{fig:plot4}

	\end{figure*} 

In the scenario without obstacles (Fig.~\ref{fig:plot3}), similar to the mobile robot experiments, E-ARC utilizes the same database for all configurations with 2, 4, 8, and 16 robots, requiring a construction time of 19.12 seconds. In contrast, Lightning struggles with scalability, as its database construction time increases from 8.4 seconds for 2 robots to 6,508.24 seconds for 16 robots. E-ARC benefits from resolving conflicts using precomputed subproblem solutions, resulting in faster planning times. For example, with 16 robots, E-ARC completes planning in 6.17 seconds, while Lightning requires 13.72 seconds.

Both E-ARC and ARC achieve a 100\% success rate across all trials, but E-ARC offers better planning times by leveraging its database of precomputed solutions.

In the scenario with obstacles (Fig.~\ref{fig:plot4}), the added environmental complexity challenges all methods. Lightning's performance deteriorates significantly, achieving only 58\% success for 2 robots and failing entirely in configurations with 16 robots. E-ARC and ARC, however, demonstrate greater resilience, achieving success rates of 100\% for all scenarios, significantly outperforming Lightning. Despite ARC maintaining the same success rate as E-ARC, it incurs longer planning times, such as 9.12 seconds for 16 robots.

\section{Conclusions}
In this paper, we introduced E-ARC, a novel extension of our previous work, ARC, by integrating experience-based planning into its conflict resolution capabilities. E-ARC provides a practical and scalable approach for incorporating experience-based techniques into MRMP, addressing the challenges that previous methods face when managing complex scenarios involving many robots.

A key contribution of E-ARC is its ability to efficiently reuse precomputed solutions. By focusing on local conflicts and smaller robot subgroups, E-ARC avoids the need for an exponentially growing database, which would otherwise be required to cover the entire multi-robot configuration space. Building on the ARC framework, E-ARC identifies relevant subproblems and applies experience-based solutions before resorting to traditional planning methods.

Our experiments demonstrate that E-ARC significantly enhances planning efficiency across a variety of scenarios, including those involving mobile and manipulator robots, especially as the number of robots increases. Its superior scalability allows it to maintain low planning times and high success rates, even as the system scales to larger teams. The compact database approach ensures consistent performance, effectively managing coordination challenges in environments with obstacles.

Future research could explore further optimization of the database retrieval process, refine strategies for repairing invalid paths, and extend E-ARC to handle more complex robot interactions.

 
	
	
	\bibliographystyle{ieeetr}
	\bibliography{robotics.bib}

\end{document}